\documentclass[10pt,twocolumn,letterpaper]{article}

\usepackage[pagenumbers]{cvpr} 

\usepackage{graphicx}
\usepackage{amsmath}
\usepackage{amssymb}
\usepackage{booktabs}

\usepackage{multirow}
\usepackage{xcolor}
\usepackage[pagebackref,breaklinks,colorlinks,citecolor=blue]{hyperref}

\usepackage[capitalize]{cleveref}
\crefname{section}{Sec.}{Secs.}
\Crefname{section}{Section}{Sections}
\Crefname{table}{Table}{Tables}
\crefname{table}{Tab.}{Tabs.}

\def\cvprPaperID{*****} 
\def\confName{CVPR}
\def\confYear{2023}

\begin{document}

\title{TokenFlow: Rethinking Fine-grained Cross-modal Alignment in Vision-Language Retrieval}

\author{Xiaohan Zou$^1$\thanks{Work done while Xiaohan Zou (zxh@bu.edu) was an intern at Kuaishou Technology.} , Changqiao Wu$^2$, Lele Cheng$^2$, Zhongyuan Wang$^2$\\
$^1$Boston University, $^2$Kuaishou Technology
}

\maketitle

\begin{abstract}
  Most existing methods in vision-language retrieval match two modalities by either comparing their global feature vectors which misses sufficient information and lacks interpretability, detecting objects in images or videos and aligning the text with fine-grained features which relies on complicated model designs, or modeling fine-grained interaction via cross-attention upon visual and textual tokens which suffers from inferior efficiency. To address these limitations, some recent works simply aggregate the token-wise similarities to achieve fine-grained alignment, but they lack intuitive explanations as well as neglect the relationships between token-level features and global representations with high-level semantics. In this work, we rethink fine-grained cross-modal alignment and devise a new model-agnostic formulation for it. We additionally demystify the recent popular works and subsume them into our scheme. Furthermore, inspired by optimal transport theory, we introduce \emph{TokenFlow}, an instantiation of the proposed scheme. By modifying only the similarity function, the performance of our method is comparable to the SoTA algorithms with heavy model designs on major video-text retrieval benchmarks. The visualization further indicates that \emph{TokenFlow} successfully leverages the fine-grained information and achieves better interpretability.
  \end{abstract}

\section{Introduction}

\begin{figure}

\centering
\includegraphics[width=0.34\textwidth]{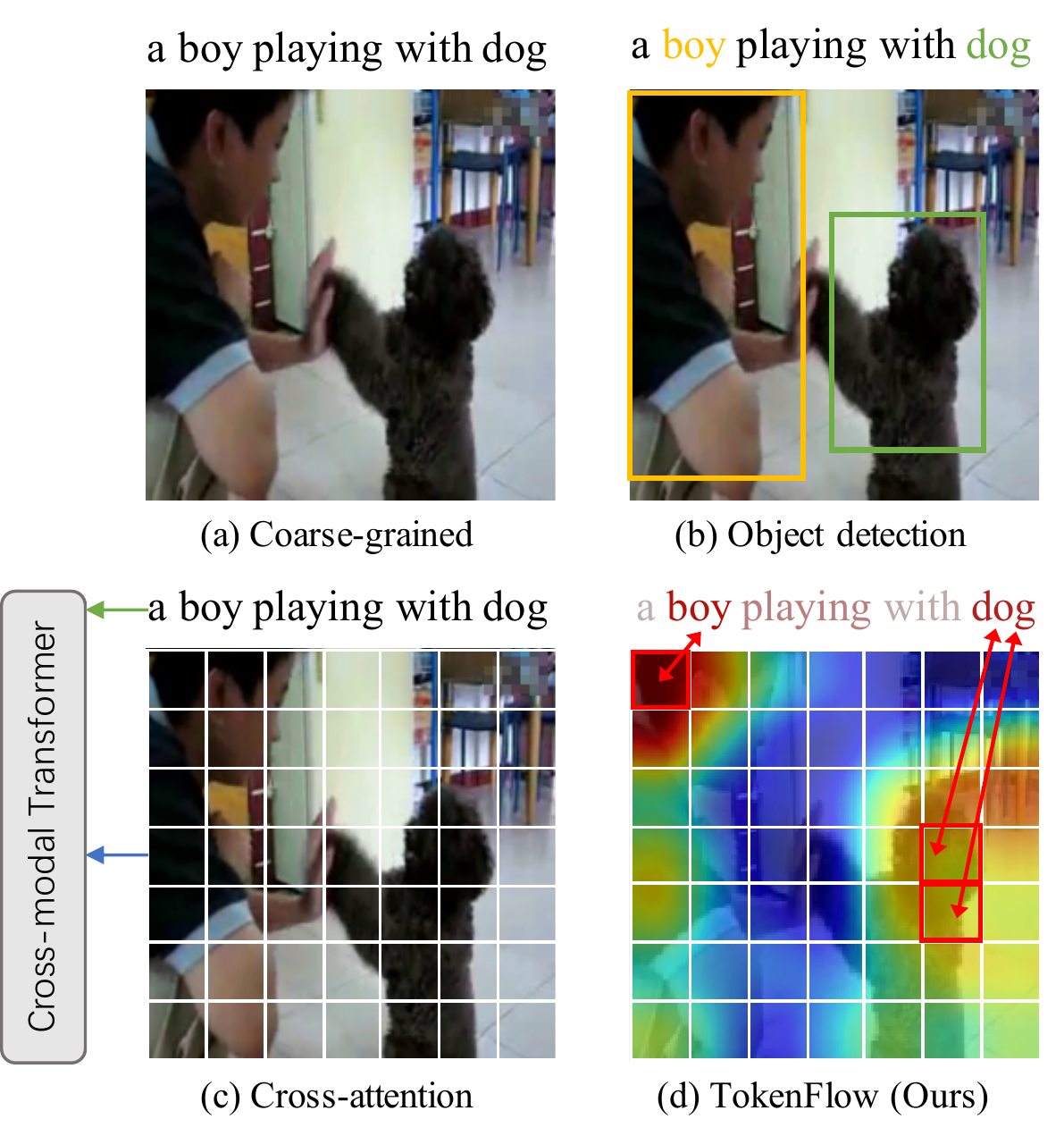}

\caption{A comparison of (a) the coarse-grained methods aligning global representations, (b) the methods relying on object detectors, (c) the methods using cross-attention layers for cross-modal interaction, and (d) our \emph{TokenFlow}.}

\label{fig:compare}

\end{figure}

Cross-modal retrieval between images (or videos) and text has become a fundamental downstream task for vision-language understanding, which aims at searching the semantic similar images or videos for a given textual query. With the rapid emergence of multimedia data on the internet, vision-language retrieval has attracted increasing attention and brought great challenges, since both visual media and text contain rich and structured details.

A variety of methods have been proposed and have shown strong superiority in learning similarities between generalizable visual and textual representations across many benchmarks. The main idea of them is to encode visual and textual inputs into the shared feature space, followed by the cross-modal alignment with global features \cite{yu2018joint, bain2021frozen, luo2021clip4clip}. Despite the superior performance in matching visual and textual features, such a line of work lacks the ability to leverage fine-level information like the relationship between visual objects and textual words.

To model fine-grained cross-modal interaction, existing methods fall into three kinds of approaches, as illustrated in Figure \ref{fig:compare}. 1) Some of them utilize pre-trained object detectors to extract region-based visual features and then fuse them with text embeddings for cross-modal training \cite{li2020oscar, gan2020large, chen2020uniter}. These works usually suffer from time-consuming region features extracting stage and require complicated architecture designs and training processes. Moreover, their ability may be limited when the object detection model fails to capture certain important information in the downstream tasks. 2) Some other works investigate fine-grained cross-modal interaction methods based on different attention mechanisms, to align the semantic space between token-wise or patch-wise representations from both modalities \cite{li2021align, kim2021vilt}. This line of work usually requires the cross-attention to be performed in an encoder-decoder architecture in both the training and inference stage and thus becomes less efficient in practice. 3) A few works achieve fine-grained cross-modal interaction by leveraging token-wise or region-word similarities in the contrastive loss, instead of using cross-attention \cite{lee2018stacked, messina2021fine, yao2021filip}.
Although these methods based on fine-level similarities are shown to be  capable of learning fine-grained representations, they directly drop the global representations that contain sufficient information, and it makes them hard to be adapted to downstream tasks with methods that use a pre-trained transformer backbones like CLIP \cite{radford2021learning}, of which the pre-training objective is aligning the global representations (classification tokens) rather than patch representations. They also neglect the fact that the relationships between single token and global statistics are also related to overall similarity. Moreover, these approaches are described in a less intuitive way which doesn’t clearly explain how they work.

In this paper, we firstly rethink the cross-modal fine-grained alignment and introduce a universal formulation for it. Then, we subsume the recent popular works which learn fine-grained interaction through token-wise or region-word similarities into our scheme and explain how they work in a clearer way. Furthermore, based on the proposed scheme, we try to model the matching problem as an optimal transport problem and define the distance of two modalities as the Earth Mover's Distance (EMD) \cite{rubner2000earth} between their structured representations. Specifically, we use spatial cross-correlation between an image (or a video) and a text as the marginal distributions when computing the optimal transport plan, where elements with larger weights generate more matching flows and thus contribute more to the overall similarity, which alleviates the issues mentioned above. However, the optimal transport problem is complicated and not friendly with time and memory consumption. Moreover, most of the existing EMD implementations don’t guarantee correctness and convergence, thus hurt the model performance. 

To address the aforementioned issues, inspired by optimal transport theory, we present \emph{TokenFlow}, a more efficient and effective instantiation of the proposed scheme which achieves promising performance. \emph{TokenFlow} computes a matching flow between the token-level features and decomposes overall similarity into several token-wise similarities with different contributions. \emph{TokenFlow} develops a very simple aligning mechanism, built of simple dot products and summations without including complex object detectors or cross-attention layers. We conduct extensive experiments to compare with other instantiations on multiple benchmarks to demonstrate the effectiveness of our algorithm.

Our main contributions are summarized
as follows:

\begin{itemize}
    \item We introduce a new perspective of fine-grained cross-modal alignment with a model-agnostic formulation.
    \item We subsume the recent popular works into our formulation and demystify them in a clearer way.
    \item We propose \emph{TokenFlow}, a novel fine-grained alignment function. Experimental results show that by learning fine-grained alignment, the performance of \emph{TokenFlow} is comparable to the SoTA algorithms with heavy model designs by only altering the similarity function, on major video-text retrieval benchmarks. Visualizations further illustrate that \emph{TokenFlow} learns meaningful fine-grained representations with promising matching ability.
\end{itemize}

\section{Related Work}

\subsection{Vision-Language Retrieval}

Existing representative works on vision-language retrieval follow the trend of learning a joint embedding space to measure the distance between visual and textual representations, which can be divided into two categories: coarse-grained and fine-grained.

Coarse-grained methods typically encode images \cite{dou2021empirical, li2021align, jia2021scaling} or videos \cite{lei2021less, luo2021clip4clip, fang2021clip2video, cheng2021improving} and textual queries to global features and accordingly map them into a common latent space, where the similarity can be measured directly with ranking loss variants. For video-text retrieval, recent methods based on pre-trained transformer CLIP \cite{radford2021learning} have achieved noticeable results and drawn increasing attention. CLIP4Clip \cite{luo2021clip4clip} is the first to apply CLIP for video-text retrieval which also proposes three different ways of aggregating video frames. CLIP2Video \cite{fang2021clip2video} captures temporal relationships of video frames and re-aligns the tokens of video clips and phrases. CAMoE \cite{cheng2021improving} extracts multi-perspective video representations including action, entity, and scene.

Extremely summarized global visual and textual descriptions may lose a lot of useful fine-grained information. For these reasons, many works try to utilize fine-level features and achieve fine-grained alignments between modalities. One line of work relies on objection detection to represent the visual input by dozens of object-centric features and then combines them with the paired text as the input of the model \cite{li2020oscar, gan2020large, chen2020uniter}. Another line of work utilizes a bunch of cross-modal transformers to learn fine-grained interaction between token-wise representations of two modalities \cite{chen2020uniter, kim2021vilt}. These methods
either require a pre-trained object detector to perform time-consuming region features extracting or cross-modal transformer layers to align the features, which significantly hinders their efficiency and scalability. In contrast, we employ a simple but effective way to align the representations of two modalities via token-level similarity matrices.

\subsection{Token-Wise/Region-Word Cross-modal Alignment}

Some efforts have been made to learn fine-grained cross-modal interaction between two modalities by leveraging token-wise or region-word similarities in the contrastive loss. TERAN \cite{messina2021fine} detects and encodes image regions at the object level with Faster-RCNN \cite{ren2015faster} and sums the maximum of the region-word similarity scores with respect to each word or region. Similar to TERAN, FILIP \cite{yao2021filip} also aggregates the maximum token-wise similarity scores according to every single feature, but it tries to directly localize fine-grained objects from visual patches, instead of using object detectors. SCAN \cite{lee2018stacked} attends differentially to important words or regions. All these works drop the global representations that contain sufficient information and neglect the relationships between fine-level features and global statistics.

\section{Approach}

In this section, we first revisit CLIP, which we use as the backbone, and denote the contrastive loss when applying it to vision-language retrieval tasks (section \ref{sec:clip}). Next, we put forward a new universal scheme for fine-level similarities based cross-modal fine-grained alignment and rethink the recent popular works (section \ref{sec:scheme}). Then, we present \emph{TokenFlow}, an instantiation of the proposed scheme inspired by optimal transport theory (section \ref{sec:proposed}). Finally, we introduce momentum distillation \cite{li2021align} which will further improve the model performance (\ref{sec:distill}).

\begin{figure*}
\centering
\includegraphics[width=\textwidth]{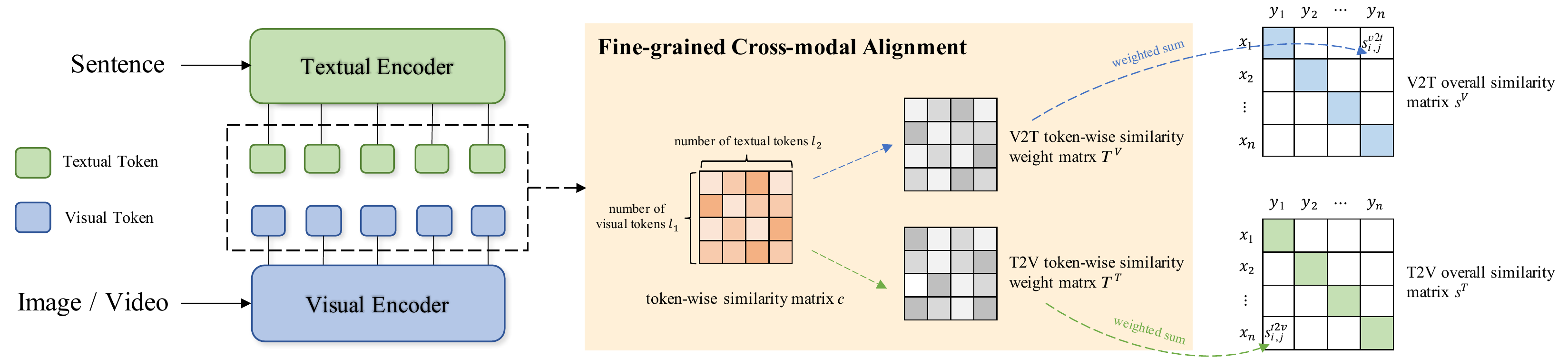}
\caption{Overall architecture of the proposed fine-grained cross-modal alignment formulation. Features of the visual and textual tokens are projected to the cross-modal joint space on top of the visual and textual encoders. Then token-wise similarities are computed and assigned with different contributions to aggregate the overall similarity.}
\label{fig:overall}
\end{figure*}

\begin{figure*}
\centering
\includegraphics[width=\textwidth]{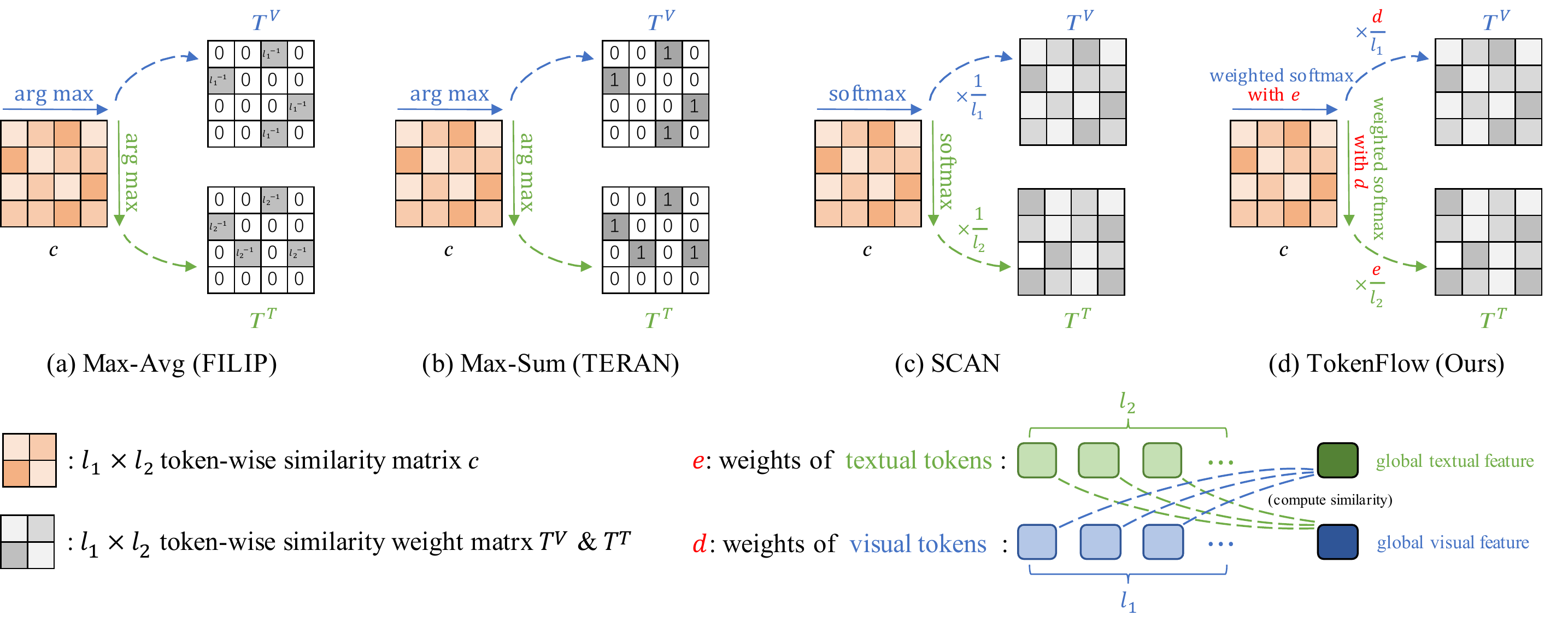}
\caption{A comparison of strategies of generating token-wise similarity weight matrix $T^T$ \& $T^V$ of different methods. (a), (b) and (c) only model the relationships between token-level (or region-word) features, while ours (d) also characterizes the importance of each token and takes relationships between token-level features and the global representations into consideration.}
\label{fig:align}
\end{figure*}

\subsection{Revisiting CLIP} \label{sec:clip}

Contrastive Language-Image Pre-training (CLIP) \cite{radford2021learning} adopts a dual-encoder framework. CLIP consists of a visual encoder $f_\theta$ based on CNN or Vision Transformer (ViT) and a text encoder $g_\phi$ based on Transformer, with a cross-modal interaction at the top. It projects visual and textual features into a joint embedding space and performs L2 normalization on them before alignment.

We consider directly fine-tuning it on the image/video-text retrieval benchmarks (without additional vision-language pre-training). Formally, given a batch of images (or videos) $\mathcal{V} = \{ v_i \}_{i=1}^N$ and the corresponding text $\mathcal{T} = \{ t_j \}_{j=1}^N$, our objective is to infer the similarity between the whole image (or video) and sentence. Under a certain similarity metric, the encoded representations $f_\theta(v_i)$ and $g_\phi(t_j)$ should be close if they are related and far apart if not. The visual-to-textual loss $L_v$ can then be formulated as:

\begin{equation}
    p^V(i) = \frac{ \exp (s_{i,i}^V ) }{\sum_{j=1}^N \exp ( s_{i,j}^V )}
\end{equation}

\begin{equation}
    L_V = \mathbb{E}_{v_i \thicksim \mathcal{V}} \big [ H (p^V(i), y^V(i)) \big ] \label{eq:pre-loss-v2t}
\end{equation}

where $s_{i,j}^{v}$ denotes the similarity of the $i$-th visual input $v_i$ to the $j$-th text $t_j$, $y^V(i)$ is the ground-truth binary label that positive pairs and negative pairs are 1 and 0 respectively, and $H(\cdot, \cdot)$ is cross-entropy formulation. Similarly, the textual-to-visual loss $L_T$ is:

\begin{equation}
    p^T(i) = \frac{ \exp (s_{i,i}^T) }{\sum_{j=1}^N \exp (s_{j,i}^T)}
\end{equation}

\begin{equation}
    L_T = \mathbb{E}_{t_i \thicksim \mathcal{T}} \big [ H (p^T(i), y^T(i)) \big ] \label{eq:pre-loss-t2v}
\end{equation}

Note that $s^V$ in Equation \ref{eq:pre-loss-v2t} does not necessarily equal $s^T$ in Equation \ref{eq:pre-loss-t2v}. The total loss of one batch is:

\begin{equation}
    L = \frac{1}{2} (L_V + L_T)
\end{equation}

The simplest method of cross-modal retrieval uses the classification token as the global feature of each image (or video) and text \cite{radford2021learning, luo2021clip4clip}, which are $f_\theta(v_i) \in \mathbb{R}^d$ and $g_\phi(t_j) \in \mathbb{R}^d$, and computes these two similarities as:

\begin{equation} \label{eq:global-sim}
    s_{i,j}^V = s_{i,j}^T = f_\theta(v_i)^\top g_\phi(t_j)
\end{equation}

The abstraction of global features loses the structures of the input during the embedding process and neglects the fine-grained alignments between the two modalities. Such problems affect the retrieval performance and also result in lacking interpretability, where the model only tells us whether the image (video) and sentence can be matched but cannot indicate what leads to the result. Therefore, in the following sections, we'll introduce how to make use of the token-level features (patch representations) of CLIP.

\subsection{Rethinking Fine-grained Cross-modal Alignment} \label{sec:scheme}

Toward a fine-grained and more interpretable cross-modal aligning mechanism, we propose a new formulation of token-wise cross-modal interaction for CLIP and other dual-encoder transformer-based backbones, as shown in Figure \ref{fig:overall}. We denote $l_1$ and $l_2$ as the number of  non-padded tokens of the $i$-th video and $j$-th text, and the corresponding representations are $\mu^i = f_\theta(v_i) \in \mathbb{R}^{l_1 \times d}$ and $\omega^j = g_\phi(t_j) \in \mathbb{R}^{l_2 \times d}$. We can then obtain a token-wise similarity matrix with:

\begin{equation}
    c^{i,j}_{s,t} = ( \mu^i_s )^\top \omega^j_t
\end{equation}

where $\mu^i_s \in \mathbb{R}^d$ is the $s$-th vector of the visual feature and $\omega^j_t \in \mathbb{R}^d$ is the $t$-th vector of the textual representation.

To formulate the fine-grained similarity of $i$-th visual input to $j$-th text, an intuitive and reasonable idea is to calculate the token-wise similarity weight matrix $T^V_{i,j}$ and define a weighted combination of token-wise similarities:

\begin{equation}
    s_{i,j}^V = \sum_{s=1}^{l_1} \sum_{t=1}^{l_2} c^{i,j}_{s,t} \big [T^V_{i,j} \big]_{s,t} \label{eq:fine-grained-formula}
\end{equation}

$T^V_{i,j}$ can be viewed as a fine-grained aligning plan for visual and textual inputs, where $\big [T^V_{i,j} \big]_{s,t}$ represents the contribution of token pair $(\mu^i_s, \omega^j_t)$ to the overall visual-to-textual alignment. Similarly, the similarity of the $j$-th text to the $i$-th visual input is:

\begin{equation}
    s_{i,j}^T = \sum_{s=1}^{l_1} \sum_{t=1}^{l_2} c^{i,j}_{s,t} \big [T^T_{i,j} \big ]_{s,t}
\end{equation}

By applying such aligning scheme, the model takes token-wise similarities into consideration and learns a fine-grained alignment between visual and textual representations. Then the next problem is the definition of the token-wise similarity weight matrix $T^V_{i,j}$ and $T^T_{i,j}$. Here we discuss several solutions used in previous works and show that the proposed scheme is a generalized version of them. We also illustrate them in Figure \ref{fig:align}. We hope this could help to understand these works in a more intuitive way.

\subsubsection{Uniform}

One trivial solution is to initialize token-wise similarity weight matrices with uniform distributions:

\begin{equation}
    T^V_{s,t} = T^T_{s,t} = \frac{1}{l_1 l_2}, \forall 1 \leq s \leq l_1, 1 \leq t \leq l_2
    \label{eq:uniform}
\end{equation}

which indicates the similarity of each token pair has an identical weight to the overall similarity. We omit the symbols $i$ and $j$ for simplicity. This solution is equal to computing the similarity with the averaged token representations of each modality as mentioned in \cite{yao2021filip}.

\subsubsection{Learnable} 

Another reasonable way is to ensemble the token-wise similarities with a learnable token-wise similarity weight matrix \cite{yao2021filip}. We empirically find that the performance of this method is not promising, refer to section \ref{sec:ablation} for more details.

\subsubsection{SCAN} 

SCAN \cite{lee2018stacked} considers fine-grained alignments by attending differentially to important words or image regions, which we'll explain in detail in Appendix \ref{appendix:scan}. The fine-grained similarity defined by SCAN can be written as:

\begin{equation}
    s_{i,j}^V = \sum_{s=1}^{l_1} \sum_{t=1}^{l_2} c_{s,t} \frac{\beta_{s,t}^V}{l_1}
\end{equation}

where $\beta_{s,t}^V$ is $c_{s,t}$ after softmax normalization \cite{chorowski2015attention}. We provide the derivation in the Appendix \ref{appendix:scan}. In this formula, SCAN can be viewed as using $\beta_{s,t}^V / l_1$ as the token-wise similarity weight matrix $T^V$, which indicates that the larger the token pair similarity is, the more it contributes to the overall alignment.

\subsubsection{Max-Avg / Max-Sum} 

FILIP \cite{yao2021filip} and TERAN \cite{messina2021fine} consider fine-grained late interaction for vision-language learning. For the $s$-th visual token (or region) $\mu_s$, they compute its similarities with all words and keep the largest one as its token-wise maximum similarity with $\omega$. They then average or sum all token-wise maximum similarities as the similarity of an image to a text. The aligning mechanism of FILIP and TERAN can also be written in the formulation of Equation \ref{eq:fine-grained-formula}:

\begin{equation}
    T^V_{s,t} =
    \begin{cases}
        1 \text{ or } \frac{1}{l_1}  &\text{if } t = \arg \max_{1 \leq r \leq l_2} c_{s,r} \\
        0 &\text{if } t \not = \arg \max_{1 \leq r \leq l_2} c_{s,r}
    \end{cases}
\end{equation}

Max-Sum and Max-Avg can be viewed as extreme versions of SCAN's aligning mechanism, where only token pairs with the largest similarity contribute to the overall alignment. However, our experiments show that these interaction mechanisms don't work on downstream vision-language retrieval tasks when using pre-trained transformer backbones like CLIP, of which the pre-training objective is aligning the global representations (classification tokens) rather than patch representations. This kind of pre-trained transformers doesn't have meaningful patch representations as the classification loss is only built on the classification tokens. Therefore, the performance of the fine-grained alignment that totally relies on the patch representations would be significantly hindered, even after being fine-tuned on the downstream tasks. Furthermore, although SCAN, FILIP, and TERAN all agree that token pairs which are more similar to each other should contribute more to the overall matching, they ignore the relationships between a single token and a global representation with high-level semantics.


\subsection{Proposed Approaches} \label{sec:proposed}

\subsubsection{Earth Mover's Distance} 

Earth Mover's Distance (EMD), which aims to seek the minimal-cost transport plan between two sets of weighted vectors, has been shown to be effective when solving the matching problem \cite{zhao2021towards, zhang2020deepemd, sarlin2020superglue}. We try to use the optimal matching flow between two sets of token vectors as the token-wise similarity weight matrix $T^V$ and $T^T$:

\begin{equation}
\begin{split}
    T^V = T^T = \arg \min_{T} \sum_{s=1}^{l_1} \sum_{t=1}^{l_2} (1 - c_{s,t}) T_{s,t} \\
    \text{subject to } T_{s,t} > 0, \sum_{s=1}^{l_1} T_{s,t} = d_s, \sum_{t=1}^{l_2} T_{s,t} = e_t
\end{split}
\end{equation}

where $d_s$ and $e_t$ are the weights of token vectors $\mu_s$ and $\omega_t$, respectively, which characterize the importance and control the total matching flows generated by each token. $(1 - c_{s,t})$ is the distance between $\mu_s$ and $\omega_t$. Here we slide the global feature of the visual (or textual) input on the token-level features of the textual (or visual) input and calculate similarity with each token:

\begin{equation}
    d_s = \mu_s^\top \overline{\omega}, \; e_t = \overline{\mu}^\top \omega_t, \label{eq:token-weight}
\end{equation}

where $\overline{\mu}$ is the global visual feature and $\overline{\omega}$ is the global textual feature. In this way, we reflect the importance of each token by comparing how close it is to the global feature of the other modality. It takes global features into consideration and thus leverages the sufficient information contained in the classification token of transformers.

However, most of the existing EMD implementations don't guarantee correctness and convergence and thus affect the performance. Meanwhile, they are complex and inefficient with respect to both time and memory.

\subsubsection{TokenFlow} 

Finally, inspired by the matching flow computed in the optimal transport problem, we introduce \emph{TokenFlow}, another instantiation of the proposed fine-grained cross-modal aligning scheme. \emph{TokenFlow} also computes a matching flow between two sets of token-level features. It decomposes the overall similarity into several weighted token-wise similarities while considering the fact that the importance of a single token also depends on the similarity with the global representations. However, compared to EMD which requires an extra optimal transport algorithm, \emph{TokenFlow} is much simpler, more efficient, and achieves better performance.

In \emph{TokenFlow}, the matching flow $T^V$ and $T^T$ are computed as:

\begin{gather}
    T^V_{s,t} = d_s \frac{\text{exp} (\lambda e_t c_{s,t})}{l_1 \sum^{l_2}_{t=1} \text{exp} (\lambda e_t c_{s,t})} \\
    T^T_{s,t} = e_t \frac{\text{exp} (\lambda d_s c_{s,t})}{l_2 \sum^{l_1}_{s=1} \text{exp} (\lambda d_s c_{s,t})}, \label{eq:tokenflow}
\end{gather}

\vspace{8pt}

As we can see in Eqation \ref{eq:tokenflow}, the weight of each token pair $\mu_s$ and $\omega_t$ is not only depend on their similarity $c_{s,t}$ but also depend on how close is $\mu_s$ or $\omega_t$ to the global representation $\overline{\omega}$ and $\overline{\mu}$.

It should be noted that we still compute the global similarity mentioned in Equation \ref{eq:global-sim} and add it to the final loss. According to Equation \ref{eq:token-weight}, learning good global representations is also very important for calculating our \emph{TokenFlow} similarities, since we need to model the relationships between them and the token-level features. In another word, the final overall similarity matrix is the weighted sum of the global and \emph{TokenFlow} similarities.

\begin{table*}[]


\centering

\begin{tabular}{ccccccccccc}

\toprule

\multirow{2}{*}{Methods} & \multicolumn{5}{c}{Text $\rightarrow$ Video} & \multicolumn{5}{c}{Video $\rightarrow$ Text} \\

\cmidrule(r){2-6} \cmidrule(r){7-11}

&  R@1      &  R@5   &   R@10   &   MdR   &   MnR
&  R@1      &  R@5   &   R@10   &   MdR   &   MnR  \\

\midrule

\color{gray}{JSFusion} \cite{yu2018joint} &\color{gray}{10.2}    &\color{gray}{31.2}    &\color{gray}{43.2}    &\color{gray}{13.0}    &\color{gray}{-}       &\color{gray}{-}       &\color{gray}{-}       &\color{gray}{-}       &\color{gray}{-}       &\color{gray}{-}     \\

\color{gray}{CE} \cite{liu2019use} &\color{gray}{20.9}    &\color{gray}{48.8}    &\color{gray}{62.4}    &\color{gray}{6.0 }    &\color{gray}{28.2}    &\color{gray}{20.6}    &\color{gray}{50.3}    &\color{gray}{64.0}    &\color{gray}{5.3 }    &\color{gray}{25.1}  \\

\color{gray}{TACo} \cite{yang2021taco}            &\color{gray}{26.7}    &\color{gray}{54.5}    &\color{gray}{68.2}    &\color{gray}{4.0 }    &\color{gray}{-   }    &\color{gray}{-   }    &\color{gray}{-   }    &\color{gray}{-   }    &\color{gray}{-   }    &\color{gray}{-   }  \\

\color{gray}{MMT} \cite{gabeur2020multi}           &\color{gray}{26.6}    &\color{gray}{57.1}    &\color{gray}{69.6}    &\color{gray}{4.0 }    &\color{gray}{24.0}    &\color{gray}{27.0}    &\color{gray}{57.5}    &\color{gray}{69.7}    &\color{gray}{3.7 }    &\color{gray}{21.3}  \\

\color{gray}{FROZEN} \cite{bain2021frozen} &\color{gray}{31.0}    &\color{gray}{59.5}    &\color{gray}{70.5}    &\color{gray}{3.0 }    &\color{gray}{-   }    &\color{gray}{-   }    &\color{gray}{-   }    &\color{gray}{-   }    &\color{gray}{-   }    &\color{gray}{-}     \\

\midrule

CLIP (zero-shot) \cite{radford2021learning}          &31.2    &53.7    &64.2    &4.0     &-       &27.2    &51.7    &62.6    &5.0     &-     \\

CLIP4Clip \cite{luo2021clip4clip}      &44.5    &71.4    &81.6    &2.0     &15.3    &42.7    &70.9    &80.6    &2.0     &11.6  \\

CLIP2Video \cite{fang2021clip2video}     &45.6    &72.6    &81.7    &2.0     &14.6    &43.5    &72.3    &82.1    &2.0     &10.2  \\

CAMoE  \cite{cheng2021improving}         &44.6    &72.6    &81.8    &2.0     &13.3    &45.1    &72.4    &83.1    &2.0     &10.0  \\

\midrule

TokenFlow (Ours)           &45.1    &72.3    &81.5    &2.0     &14.8    &43.2    &72.3    &81.9    &2.0     &10.2  \\

TokenFlow + MD (Ours)      &\textbf{46.1}    &\textbf{72.7}    &\textbf{82.0}    &\textbf{2.0}     &\textbf{13.6}    &\textbf{45.4}    &\textbf{73.5}    &\textbf{83.7}    &\textbf{2.0}     &\textbf{10.0}  \\

\bottomrule

\end{tabular}

\caption{Video-text results on MSR-VTT 1K-A split. MD represents Momentum Distillation. We gray out models that don't use CLIP as the backbone for a fair comparison.}

\label{tab:msrvtt}

\end{table*}

\begin{table}

\centering

\resizebox{0.48 \textwidth}{!}{
    \begin{tabular}{ccccc}
    
    \toprule
    
    Methods         &  R@1   &  R@5   &  R@10  &  MdR  \\
    
    \midrule       
    
    \color{gray}{CE} \cite{liu2019use}     &\color{gray}{19.8}    &\color{gray}{49.0}    &\color{gray}{63.8}    &\color{gray}{6.0}  \\
    
    \color{gray}{FROZEN} \cite{bain2021frozen}    &\color{gray}{33.7}    &\color{gray}{64.7}    &\color{gray}{76.3}    &\color{gray}{3.0}  \\
    
    \midrule
    
    CLIP (zero-shot) \cite{radford2021learning}    &37.0    &64.1    &73.8    &3.0    \\
    
    CLIP4Clip \cite{luo2021clip4clip}    &46.2    &76.1    &84.6    &2.0    \\
    
    CLIP2Video \cite{fang2021clip2video}    &47.0    &\textbf{76.8}    &\textbf{85.9}    &2.0    \\
    
    CAMoE \cite{cheng2021improving}    &46.9    &76.1    &85.5    &-      \\
    
    \midrule
    
    TokenFlow (Ours)       &46.7    &76.4    &85.1    &2.0    \\
    
    TokenFlow + MD (Ours)   &\textbf{47.1}    &76.1    &85.5    &\textbf{2.0}    \\
    
    \bottomrule
    
    \end{tabular}
}

\caption{Video-text retrieval results on MSVD dataset. All metrics are measured for Text-to-Video Retrieval.}
\label{tab:msvd}

\end{table}

\begin{table}

\centering

\resizebox{0.48 \textwidth}{!}{
    \begin{tabular}{ccccc}
    
    \toprule
    
    Methods         &  R@1   &  R@5   &  R@50  &  MdR  \\
    
    \midrule
    
    \color{gray}{CE}    &\color{gray}{18.2}    &\color{gray}{47.7}    &\color{gray}{-}    &\color{gray}{6.0} \\
    
    \color{gray}{MMT}        &\color{gray}{28.7}    &\color{gray}{61.4}    &\color{gray}{94.5}    &\color{gray}{5.0} \\
    
    
    
    \color{gray}{ClipBERT}   &\color{gray}{21.3}    &\color{gray}{49.0}    &\color{gray}{-}       &\color{gray}{6.0} \\
    
    \color{gray}{TACo}       & \color{gray}{30.4}   &\color{gray}{61.2}    &\color{gray}{93.4}    &\color{gray}{3.0} \\
    
    \midrule
    
    CLIP4Clip \cite{luo2021clip4clip}    &40.5    &72.4    &98.1    &2.0     \\
    
    
    \midrule
    
    TokenFlow (Ours)      &41.1    &72.8    &98.1    &2.0    \\
    
    TokenFlow + MD (Ours)    &\textbf{41.5}    &\textbf{73.1}    &\textbf{98.8}    &\textbf{2.0}    \\
    
    \bottomrule
    
    \end{tabular}
}

\caption{Video-text retrieval results on ActivityNet dataset. All metrics are measured for Text-to-Video Retrieval.}
\label{tab:activitynet}

\end{table}

\begin{table*}[]


\centering

\begin{tabular}{ccccccc}

\toprule

\multirow{2}{*}{Methods} & \multicolumn{3}{c}{Flickr30K} & \multicolumn{3}{c}{MS-COCO} \\

\cmidrule(r){2-4} \cmidrule(r){5-7}

&  R@1      &  R@5   &   R@10
&  R@1      &  R@5   &   R@10  \\

\midrule

CLIP-ViT-B/32 (zero-shot) \cite{radford2021learning} \dag         &59.7    &84.8    &90.7    &30.5    &56.0    &66.9     \\

CLIP-ViT-B/32 (fine-tune) \cite{radford2021learning} \dag  &74.1    &93.4    &\textbf{96.7}    &44.6    &72.1    &81.6     \\

\midrule

TokenFlow (Ours)       &75.2    &93.8    &96.0    &46.2    &73.0    &82.0     \\

TokenFlow + MD (Ours)   &\textbf{75.9}    &\textbf{94.2}    &96.5    &\textbf{47.5}    &\textbf{74.1}    &\textbf{82.7}     \\

\bottomrule

\end{tabular}

\caption{Image-text retrieval results on Flickr30K and MS-COCO datasets (Karpathy split). All metrics are measured for Text-to-Image Retrieval (image retrieval given text query). $\dag$ indicates the results re-produced by us.}

\label{tab:itr}

\end{table*}

\subsection{Momentum Distillation} \label{sec:distill}

Image-text (or video-text) pairs in vision-language retrieval tasks may be weakly-correlated. That is, an image (or a video) may not cover all the words of its related caption, and a text description may not fully describe its corresponding image (or video). The issue is even more obvious in video-text retrieval, where many of the video frames have to be dropped by the frame sampling strategy. Therefore, we plant momentum distillation which tackles the weak correlation between image-text pairs introduced in ALBEF \cite{li2021align} into the vision-language retrieval task. We describe the details in Appendix \ref{appendix:md}.

\begin{table}

\centering

\begin{tabular}{ccccc}

\toprule

Methods         &  R@1   &  R@5   &  R@10  &  MdR  \\

\midrule

CLIP4Clip \cite{luo2021clip4clip}      &46.2    &76.1    &84.6    &2.0    \\

\midrule

Uniform \cite{yao2021filip}        &38.4    &66.8    &77.4    &2.0    \\

Learnable \cite{yao2021filip}   &42.7    &71.7    &81.7    &2.0      \\

Max-Avg \cite{yao2021filip}   &42.2    &71.9    &81.7    &2.0      \\

SCAN \cite{lee2018stacked}  &43.0    &72.6    &82.2    &2.0      \\

EMD             &44.5    &74.0	  &83.6	   &2.0    \\

\midrule

TokenFlow (Ours)        &\textbf{46.7}    &\textbf{76.4}    &\textbf{85.1}    &\textbf{2.0}    \\

\bottomrule

\end{tabular}

\caption{Ablation study for the different fine-grained cross-modal aligning strategy on MSVD dataset. All metrics are measured for Text-to-Video Retrieval. CLIP4Clip \cite{luo2021clip4clip} is recognized as the baseline, which doesn't employ any fine-grained interaction mechanism.}
\label{tab:ablation-msvd}

\end{table}

\begin{table}

\centering

\resizebox{0.48 \textwidth}{!}{

    \begin{tabular}{cccc}
    
    \toprule
    
    Methods         &  R@1   &  R@5   &  R@10  \\
    
    \midrule
    
    CLIP-ViT-B/32 (fine-tune) \cite{radford2021learning} \dag   &74.1    &93.4    &96.7    \\
    
    \midrule
    
    Uniform \cite{yao2021filip}    & 67.6    & 89.9    & 93.8    \\
    
    Learnable \cite{yao2021filip}    & 69.8    & 90.5    & 94.3    \\
    
    Max-Avg \cite{yao2021filip}    & 73.8    & 92.4    & 95.7    \\
    
    SCAN \cite{lee2018stacked}    &70.1    &90.5    &94.6   \\
    
    EMD    & 72.2    & 91.6    & 95.1    \\
    
    \midrule
    
    TokenFlow  (Ours)        &\textbf{75.2}    &\textbf{93.8}    &\textbf{96.0}   \\
    
    \bottomrule
    
    \end{tabular}
}

\caption{Ablation study for the different fine-grained cross-modal aligning strategy on Flickr30K dataset. All metrics are measured for Text-to-Image Retrieval. CLIP-ViT-B/32 (fine-tune) \cite{radford2021learning} is recognized as the baseline, which doesn't employ any fine-grained interaction mechanism. $\dag$ indicates the results re-produced by us.}
\label{tab:ablation-flickr30k}

\end{table}

\section{Experiments}

\subsection{Experimental Settings}

We compare our approach with the state-of-the-art methods for video-text retrieval tasks and image-text retrieval tasks. For video-text retrieval, we conduct experiments on three major benchmarks including MSVD \cite{chen2011collecting}, MSR-VTT \cite{xu2016msr}, and ActivityNet \cite{krishna2017dense}. We also evaluate the proposed method for the image-text retrieval task on Flickr30K \cite{young2014image} and MS-COCO \cite{lin2014microsoft}. See Appendix \ref{appendix:datasets} for more details regarding the datasets.

We report standard retrieval metrics: recall at rank K (R@K), median rank (MdR), and mean rank (MnR). Please see Appendix \ref{appenix:metric} for the explanation of each evaluation metric and see Appendix \ref{appendix:implementation} for implementation details.

\subsection{Comparison to the State of the Art}

\subsubsection{Video-Text Retrieval}

We compare our approach against the SoTA methods and report the results in Table \ref{tab:msrvtt} and \ref{tab:msvd}. As we can see, our method outperforms strong baseline CLIP4Clip \cite{luo2021clip4clip} on all three benchmarks. We also observe that the performance of our method which alters the aligning function only and aggregates video frames in the simplest parameter-free fashion is comparable to SoTA methods like CLIP2Video \cite{fang2021clip2video} and CAMoE \cite{cheng2021improving} with heavy model designs. It should be noted that we neither utilize complex temporal modeling strategies \cite{fang2021clip2video} nor incorporate other learnable modules \cite{fang2021clip2video, cheng2021improving}. The results indicate the effectiveness of the proposed alignment mechanism based on fine-grained features. 

Furthermore, adding momentum distillation makes our algorithm outperforms CLIP2Video and CAMoE (except on ActivityNet). The improvement brought by momentum distillation is especially significant on datasets with shorter videos like MSR-VTT. Thus we conclude that handling the weak correlation between video-text pairs is helpful in video-text retrieval tasks.

\subsubsection{Image-Text Retrieval}

Table 2 reports the results on MS-COCO and Flickr30K dataset. We neglect models that don’t use CLIP (ViT-B/32) as the backbone for a fair comparison. The performance of our method surpasses that of directly fine-tuning CLIP using a coarse-grained objective.

\begin{figure*}

\centering
\includegraphics[width=\textwidth]{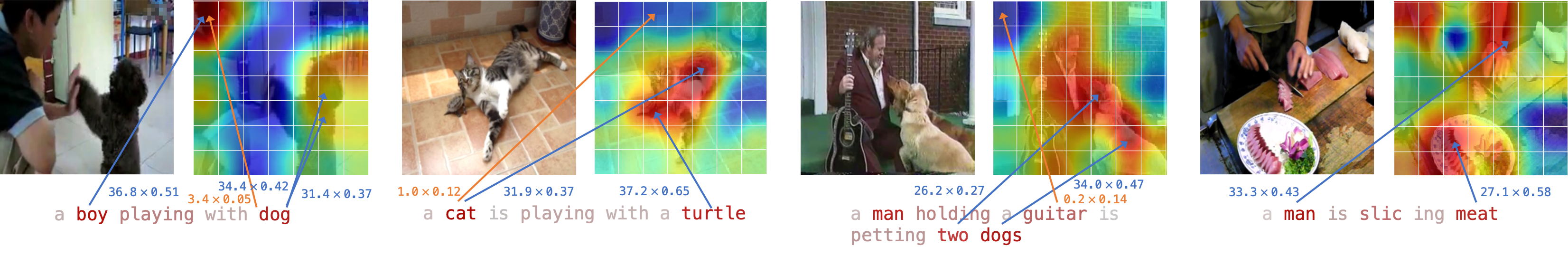}

\caption{Visualization of text-to-video fine-grained alignment on MSVD dataset. We patchify the middle (6th) frame of video to 7 $\times$ 7 image patches and use heatmaps to show the weights of each patch (Equation \ref{eq:token-weight}). We then label the weights of textual tokens (Equation \ref{eq:token-weight}) by color. We also  illustrate some of the representative word-patch similarities as well as their contributions to the overall alignment in the form of $\hat{T}^{t2v}_{s,t} \times c_{s,t}$, where $\hat{T}^{t2v}_{s,t}$ is the re-scaled matching flow and $c_{s,t}$ is the token-wise similarity between word-patch pair $(\mu_s, \omega_t)$.}

\label{fig:vis}

\end{figure*}

\subsection{Ablation Studies} \label{sec:ablation}

We conduct extensive experiments to study the effect of different instantiations of the proposed fine-grained alignment scheme on MSVD and Flickr30K dataset. See section \ref{sec:scheme} for details of each compared approach. As shown in Table \ref{tab:ablation-msvd} and Table \ref{tab:ablation-flickr30k}, \emph{TokenFlow} outperforms all of others, which indicates its effectiveness. During pretraining, CLIP's loss is only built on the classification token (visual encoder) and $<EOS>$ token (textual encoder), which leads to the lack of meaningful information on other tokens. Thus it is not surprising that \emph{Uniform}, \emph{SCAN} and \emph{Max-Avg} which only leverages token-level features perform even worse than directly fine-tuning CLIP. We then conclude that FILIP and SCAN's performance relies on informative patch or region features obtained by large-scale vision-language pre-training or a pre-trained object detector, which are both infeasible in our experiment settings. Furthermore, the results also show that \emph{SCAN} outperforms \emph{Max-Avg} on MSVD but underperforms \emph{Max-Avg} on Flickr30K. \emph{Learnable} also performs poorly, which indicates that it is hard for a model to learn the token-wise similarity weight matrix well without any other information.

\subsection{Visualization} \label{sec:visualization}

To better understand how \emph{TokenFlow} works and its capability of capturing cross-modal fine-grained relationships, we visualize token weights, token-wise similarities, and fine-grained alignment based on token-wise similarities on MSVD dataset in Figure \ref{fig:vis}, where all video-sentence pairs are positive pairs.

\paragraph{Visualization Method.}

For videos, we display visualization on the middle frame (the 6th frame, specifically). Each video frame is patchified to 7 $\times$ 7 image patches. We then visualize the weight of each patch ($d_s$ in Equation \ref{eq:token-weight}) through heatmaps. For sentences, we label the weight of each token ($e_t$ in Equation \ref{eq:token-weight}) by color, where the darker the color is, the larger the weight is. Then, we show the text-to-video matching flow $T^{t2v}_{s,t}$ and the token-wise similarity $c_{s,t}$ for some of the word-patch pairs $(\mu_s, \omega_t$). We draw arrows between the word-patch pairs that make a large (or small) contribution to the overall similarity in blue (or orange). The formula along
with an arrow takes the form of $T^{t2v}_{s,t} \times c_{s,t}$. Since each $T^{t2v}_{s,t}$ is relatively small, we use a re-scaled version $\hat{T}^{t2v}_{s,t} = 100 \; T^{t2v}_{s,t}$.

\paragraph{Observations.} Figure \ref{fig:vis} shows the visualization of fine-grained text-to-video alignment for \emph{TokenFlow} on MSVD. As can be seen, \emph{TokenFlow} effectively learns the fine-grained interaction. First, we find that the heatmaps demonstrating weights of image patches can focus on discriminative parts (e.g. boy, dog, cat.). Second, we observe that \emph{TokenFlow} assigns similar pairs a higher $T^{t2v}_{s,t}$ while enforcing a lower $T^{t2v}_{s,t}$ to the pairs that are less similar. In this way, our method aligns visual tokens with corresponding textual words correctly. Third, compared with \cite{yao2021filip} and \cite{lee2018stacked}, our method takes relationships between local features and global statistics into consideration and learns to capture the meaning of words like "two", which is hard to learn by only using token-level features. Compared with \cite{zhao2021towards} and \cite{zhang2020deepemd}, which use an optimal transport algorithm to learn the fine-grained alignment, our method learns the alignment in a much simpler and more efficient way.

\section{Conclusion}

In this paper, we rethink fine-grained cross-modal alignment in vision-language retrieval tasks and propose a universal model-agnostic scheme for it. We then discuss several recent popular works that also consider fine-grained interaction and show that the proposed scheme is a general version of them all. Our scheme provides a better perspective for understanding the fine-grained alignment mechanism. Furthermore, inspired by optimal transport theory, we introduce a new fine-grained aligning strategy called \emph{TokenFlow} which outperforms the above-mentioned methods and achieves performance that is comparable to SoTA approaches on vision-language retrieval tasks without any complex model designs.

{\small
\bibliographystyle{ieee_fullname}
\bibliography{main}
}

\appendix

\section{Stacked Cross Attention (SCAN)} \label{appendix:scan}

Stacked Cross Attention (SCAN) \cite{lee2018stacked} focuses on image-text retrieval task. For image-to-text alignment, it first attends textual tokens $\omega$ to $s$-th visual token $\mu_s$:

\begin{equation}
    a_s^V = \sum_{t=1}^{l_2} \beta_{s,t}^V \omega_t
\end{equation}

where the coefficient is:

\begin{equation}
    \beta_{s,t}^V = \frac{\text{exp} (\lambda c_{s,t})}{\sum^{l_2}_{t=1} \text{exp} (\lambda c_{s,t})}
\end{equation}

and $\lambda$ is the inversed temperature of the softmax function. It then computes the similarity between the attended text vector and each visual token and summarizes it as the similarity between image and sentence:

\begin{equation}
    s_{i,j}^V = \frac{1}{l_1} \sum_{s=1}^{l_1} \mu_s^\top a_s^V
\end{equation}

SCAN can be subsumed into the scheme proposed in Section \ref{sec:scheme} by:

\begin{align}
    s_{i,j}^V &= \frac{1}{l_1} \sum_{s=1}^{l_1} \mu_s^\top a_s^V \\
        &= \frac{1}{l_1} \sum_{s=1}^{l_1} \left ( \mu_s^\top \sum_{t=1}^{l_2} \beta_{s,t}^V \omega_t \right ) \\
        &= \sum_{s=1}^{l_1} \sum_{t=1}^{l_2} c_{s,t} \frac{\beta_{s,t}^V}{l_1}
\end{align}

\section{Momentum Distillation} \label{appendix:md}

We maintain a teacher model with identical architecture to the student and update it with an exponential-moving-average strategy for generating pseudo-targets $y^V_m$ and $y^T_m$. We also maintain four queues to store the most recent global visual representations, fine-grained visual representations, global textual representations, and fine-grained textual representations extracted by the teacher model, respectively. During training, pseudo-targets, similarities, and contrastive losses (both global- and fine-level) are not only computed within the batch but also in queues. We then combine the pseudo-targets with the ground-truth label by:

\begin{gather}
    \hat{y}^V(i) = \alpha y^V_m(i) + (1 - \alpha) y^V (i) \\
    \hat{y}^T(i) = \alpha y^T_m(i) + (1 - \alpha) y^T (i)
\end{gather}

Finally, we compute the contrastive loss by replacing $y^V (i)$ and $y^T (i)$ with $\hat{y}^V(i)$ and $\hat{y}^T(i)$ in Equation \ref{eq:pre-loss-v2t} and Equation \ref{eq:pre-loss-t2v}.

\section{Datasets} \label{appendix:datasets}

Here we list the benchmark datasets we considered. For video-text retrieval task:

\begin{itemize}
    \item \textbf{MSVD} dataset
    \cite{chen2011collecting} is composed of 1,970 videos, each with a length that ranges from 1 to 62 seconds and approximately 40 associated sentences in English. Train, validation, and test are split into 1,200, 100, and 670 videos.
    \item \textbf{MSR-VTT} dataset \cite{xu2016msr} consists of 10,000 videos, each with a length ranging from 10 to 32 seconds and 200,000 captions. We report the results on the 1K-A split in our paper, where 9,000 videos with all corresponding captions are used for training, and another 1,000 video-text pairs are used as the test set.
    \item \textbf{ActivityNet} dataset \cite{krishna2017dense} includes 20,000 YouTube videos. We follow \cite{zhang2018cross, luo2021clip4clip} to concatenate all the descriptions of a video to form a paragraph and evaluate the model with video-paragraph retrieval on the ‘val1’ split.
\end{itemize}

For image-text retrieval task:

\begin{itemize}
    \item \textbf{Flickr30K} dataset \cite{young2014image} contains 31,000 images with a total of 158,915 English sentences, where each image is annotated with around 5 sentences. We adopt the widely used Karpathy split \cite{karpathy2015deep}, where 29,000 images are used for training, 1,000 images for validation, and the remaining 1,000 images are used as the test set.
    \item \textbf{MS-COCO} dataset \cite{lin2014microsoft} has about 123,000 images and each image comes with at least 5 sentences. We also adopt the Karpathy split  \cite{karpathy2015deep} for this benchmark, where the numbers of images of training, validation, and test set are 113,287, 5,000, and 5,000, respectively.
\end{itemize}

\section{Evaluation Metrics} \label{appenix:metric}

Recall at rank K (R@K) is the percentage for which the correct result is found in the top-K retrieved points to the query sample, eg. R@1, R@5, and R@10 (or R@50 for ActivityNet) are reported in this paper. Median rank (MdR) and mean rank (MnR) calculate the median and mean of the correct results in the ranking, respectively. The higher R@K and lower MdR and MnR indicate better performance.

\section{Implementation Details} \label{appendix:implementation}

We initialize our vision and text encoder with CLIP (ViT-B/32). We don't incorporate any other learnable modules. Visual and semantic embeddings are projected to 512 dimensions before being aligned. Logit scaling $l$ is the same as CLIP. The model is finetuned with AdamW optimizer and cosine scheduler with warmup in 10 epochs on 4 NVIDIA Tesla V100 GPUs. The learning rate is set to 1e-6 on MS-COCO and 1e-7 on other datasets. The maximum number of caption tokens is 32. For momentum distillation, the parameter $\alpha$ for updating the teacher model is set as 0.95 and the lengths of the feature queues are set to 16. For the video-text retrieval task, we uniformly sample 12 frames for each video for training. We simply average features of all video frames instead of performing any other complicated temporal modeling.

\end{document}